\documentclass[sigconf,screen]{acmart}

\setcopyright{cc}
\setcctype{by}
\acmDOI{10.1145/3842650.3843171}
\acmYear{2026}
\copyrightyear{2026}
\acmISBN{979-8-4007-2968-3/2026/10}
\acmConference[EXPRESS '26]{Proceedings of the 2nd Workshop on Explainable and Reliable Software Systems}{October 4--9, 2026}{Oakland, CA, USA}
\acmBooktitle{Proceedings of the 2nd Workshop on Explainable and Reliable Software Systems (EXPRESS '26), October 4--9, 2026, Oakland, CA, USA}
\acmSubmissionID{splashws26expressmain-p4-p}
\received{2026-06-28}
\received[accepted]{2026-07-31}

\usepackage{microtype}

\usepackage{booktabs}
\usepackage{tikz}
\usetikzlibrary{positioning,calc}
\usepackage{listings}
\lstdefinestyle{paircode}{
  language=Python,
  basicstyle=\fontsize{7.5}{9}\selectfont\ttfamily,
  keywordstyle=\bfseries,
  commentstyle=\color{black!65},
  numbers=left, numberstyle=\tiny\color{black!55},
  numbersep=4pt, xleftmargin=11pt,
  columns=fullflexible, keepspaces=true,
  showstringspaces=false, aboveskip=3pt, belowskip=0pt,
  escapeinside={(*@}{@*)}
}
\newlength{\pairwidth}
\newsavebox{\workerA}
\newsavebox{\workerB}
\newsavebox{\mergeTrace}
\usepackage{xspace}
\usetikzlibrary{arrows.meta,positioning,fit,backgrounds}
\usepackage{listings}
\newcommand{\rev}[1]{{#1}}

\begin{document}

\title[Passes Alone, Fails Together]{Passes Alone, Fails Together: Benchmarking Semantic Coordination in Parallel LLM-Agent Development}

\author{Haocheng Xia}
\orcid{0000-0002-8317-6217}
\affiliation{%
  \institution{University of Illinois Urbana-Champaign}
  \city{Champaign}
  \country{USA}
}
\email{hxia7@illinois.edu}

\author{Eugene Wu}
\orcid{0000-0003-4254-6688}
\affiliation{%
  \institution{Columbia University}
  \city{New York}
  \country{USA}
}
\email{ewu@cs.columbia.edu}

\author{Yongjoo Park}
\correspondingauthor
\orcid{0000-0003-3786-6214}
\affiliation{%
  \institution{University of Illinois Urbana-Champaign}
  \city{Champaign}
  \country{USA}
}
\email{yongjoo@illinois.edu}

\begin{abstract}
\rev{Parallel coding agents can produce patches that work alone but fail when merged. This happens when one agent changes an interface or rule that another agent still relies on. We study these failures with \textsc{stale}, a benchmark for semantic coordination. Our evaluation runs the same tests on each patch alone and on their combination, counting only failures introduced by combining the patches. We use three tiers: synthetic tasks with controlled interface changes, pairs of merged pull requests, and constructed tasks that use real Django helpers. Among 834 runs on 417 mined Django pairs, only one showed interference after correcting the grading procedure. On constructed tasks using 12 Django helpers, interference occurred in 97\% of runs. A message describing the completed concurrent change recovered 82\% of runs. Reviewed pull requests may contain few unresolved parallel changes, even when agents fail on controlled tasks using real code. The constructed failure rates do not estimate how often these problems occur in practice.}
\end{abstract}


\maketitle
\section{Introduction}

Parallel multi-agent systems use different coordination models, including shared state, explicit orchestration, asynchronous updates, and isolated workers whose outputs are merged afterwards~\cite{chatdev,metagpt}. We focus on a specific scenario where workers do not receive updates from their peers while on the job. This scenario can occur as a complete workflow or as a temporary information gap within a coordinated system. 
It is efficient and harmless for independent tasks~\cite{swegym,ogenrwot2026agenticflictlargescaledatasetmerge}, but dangerous when one patch depends on an interface or invariant that is being changed concurrently. 
Each patch may then pass in isolation while their composition fails. We call the problem of maintaining consistency across such concurrent, coupled changes \emph{semantic coordination}.

\paragraph{Motivating Example.} 
Figure~\ref{fig:semantic-coordination} illustrates how independently correct patches can become inconsistent when combined. Two workers start from a service that reads and writes prices directly in a database. Worker A adds caching and modifies the existing update function to invalidate cached prices. Without seeing A's changes, Worker B adds a bulk-update function that writes directly to the database. Then a query returns 100 even after the database price changes to 80. The failure arises because A introduces an invariant that B's new write path violates: every price update must invalidate the corresponding cache. Keeping this invariant across patches requires semantic coordination beyond conflict-free code merging.

\begin{figure}[t]
\fontsize{8}{9.5}\selectfont
\centering
\setlength{\pairwidth}{\columnwidth}
\begin{lrbox}{\workerA}
\begin{minipage}[t]{0.455\pairwidth}
\textbf{(a) A: add caching}
\begin{lstlisting}[style=paircode]
cache = {}

def get_price(k):
    if k not in cache:
        cache[k] = db[k]
    return cache[k]

def update_price(k, v):
    db[k] = v
    (*@\colorbox{black!10}{\strut\texttt{cache.pop(k, None)}}@*)
\end{lstlisting}
\end{minipage}
\end{lrbox}
\begin{lrbox}{\workerB}
\begin{minipage}[t]{0.455\pairwidth}
\textbf{(b) B: add bulk updates}
\begin{lstlisting}[style=paircode]
def bulk_update(prices):
    (*@\colorbox{black!10}{\strut\texttt{db.update(prices)}}@*)
\end{lstlisting}
\medskip
{\fontsize{8}{10}\selectfont\raggedright\itshape
New write path bypasses the cache added by A.\par}
\end{minipage}
\end{lrbox}
\begin{lrbox}{\mergeTrace}
\begin{minipage}{0.96\pairwidth}
\textbf{(c) Merged execution}\\[-4pt]
\begin{lstlisting}[style=paircode]
# Initially: db = {"x": 100}; cache = {}
get_price("x")          # 100; caches price
bulk_update({"x": 80})  # database only
get_price("x")          # 100; expected 80
\end{lstlisting}
\end{minipage}
\end{lrbox}
\begin{tikzpicture}[
  panel/.style={anchor=north west,inner sep=0pt,outer sep=0pt},
  rule/.style={draw=black!55,line width=0.4pt}
]
\node[panel] (a) {\usebox{\workerA}};
\node[panel,right=0.045\pairwidth of a.north east,anchor=north west]
  (b) {\usebox{\workerB}};
\draw[rule] ([xshift=0.0225\pairwidth]a.north east) --
  ([xshift=0.0225\pairwidth]a.south east);
\node[panel,below=13pt of a.south west,anchor=north west]
  (trace) {\usebox{\mergeTrace}};
\draw[rule] ([yshift=6pt]trace.north west) --
  ([yshift=6pt]trace.north east);
\end{tikzpicture}
\caption{Independently correct patches merge without textual conflicts but produce stale prices.}
\Description{Worker A adds a price cache and invalidation to the existing update function. Worker B adds bulk updates that bypass the cache. After merging, a bulk update changes a database price from 100 to 80, but the cached query still returns 100.}
\label{fig:semantic-coordination}
\end{figure}

\rev{Version control helps developers combine parallel changes, but a clean merge does not guarantee correct behavior. For example, one patch may rename a helper that another patch still calls, add a required argument that another caller omits, or change a rule that other code relies on. These inconsistencies may appear only when tests run on the combined code. By then, diagnosing the failure requires understanding both patches.}

\rev{A clean merge can still fail, even when agents edit different files. AutoGen~\cite{autogen} and MetaGPT~\cite{metagpt} coordinate agents through roles and messages, as do ChatDev~\cite{chatdev} and Agent Teams~\cite{claude-agent-teams}. We measure failures caused by missing updates and test how many an agent can avoid when it sees the concurrent change.}

\begin{figure}[ht]
\centering
\footnotesize
\begin{tikzpicture}[
  c/.style={draw,minimum height=6.2mm,font=\small,align=center,inner sep=2pt},
  neu/.style={c,minimum width=20mm,fill=black!5},
  pass/.style={c,minimum width=25mm,fill=green!13},
  fail/.style={c,minimum width=25mm,fill=red!14},
  hdr/.style={font=\small\bfseries},
  rl/.style={font=\small,anchor=east}]
\node[hdr] at (2.2,0.72) {\texttt{git} merge};
\node[hdr] at (4.5,0.72) {behavioral tests};
\node[rl]   at (0.95,0)    {change $A$ alone};
\node[neu]  at (2.2,0)     {--};
\node[pass] at (4.5,0)     {pass};
\node[rl]   at (0.95,-0.7) {change $B$ alone};
\node[neu]  at (2.2,-0.7)  {--};
\node[pass] at (4.5,-0.7)  {pass};
\node[rl]   at (0.95,-1.4) {$A\parallel B$\,(\textsc{blind})};
\node[neu]  at (2.2,-1.4)  {no conflict};
\node[fail] at (4.5,-1.4)  {\textbf{fail}\,$=\Delta_{\mathrm{blind}}$};
\node[rl]   at (0.95,-2.1) {$A\!\to\!B$\,(\textsc{informed})};
\node[neu]  at (2.2,-2.1)  {no conflict};
\node[pass] at (4.5,-2.1)  {pass};
\end{tikzpicture}
\caption{A clean merge can still introduce test failures. We count a test in $\Delta_{\mathrm{blind}}$ only if it passes with each patch alone and fails with the patches combined. An informed agent sees the other worker's completed edit before solving its task. Comparing these conditions estimates how much complete information can help.}
\Description{Comparison of patches alone, patches merged without coordination, and sequential informed work. Individual patches pass; a clean blind merge introduces failures; an informed agent can adapt to the other patch.}
\label{fig:pipeline}
\end{figure}

\rev{A \emph{blind} agent works without seeing the other worker's changes; an \emph{informed} agent works afterward and sees the completed edit. We call the informed setting an \emph{oracle} because it provides complete information that an agent working in parallel would usually lack. This comparison estimates \emph{semantic coordination headroom}: how much failure coordination could prevent under ideal visibility. To measure interference between fixed patches, we count tests that pass with each patch alone but fail with both patches together. We denote this count by $\Delta_{\mathrm{blind}}$ and also report whether it is greater than zero. These measures describe failures detected by the tests; they do not measure defect severity or how often conflicts occur in practice. We study three types of instances because historical pull requests may already have resolved the conflicts we want to measure.}

\rev{We make three contributions. \textbf{(1)} We define semantic coordination headroom and a grading procedure that tests the same patches, both alone and together, using the same test set. This separates failures caused by patch interaction from differences between agent runs or grading conditions. \textbf{(2)} We provide a deterministic pipeline for finding related pull requests and checking that both tasks can be evaluated from a shared base commit. It produces 447 pairs across Django, SymPy, xarray, and seaborn. Experiments on the 417 Django pairs find almost no interference after correcting the grading procedure. \textbf{(3)} We demonstrate the failure mechanism in controlled experiments: synthetic tasks vary how many interfaces change, and 36 constructed tasks apply these changes to 12 real Django helpers. The controlled tasks isolate failures caused by stale information. The code is available in the benchmark repository.\footnote{\url{https://github.com/illinoisdata/STALE-bench}}}

\section{Benchmark Design}
\label{sec:design}

\paragraph{Tasks and Evaluation Settings.}
In the mined tier, each instance contains $N$ related pull requests (PRs), $\{p_i\}_{i=1}^N$, evaluated from a shared base commit $b$. Each PR provides a task description, a human reference solution, and a hidden \emph{gold test patch}. The task description comes from the issue or PR, with text explaining the solution removed. The gold tests encode the expected behavior: they fail before the fix and pass afterward (fail-to-pass tests, $\mathrm{F2P}_i$).
We define three evaluation settings:
\begin{itemize}
\item \rev{\textsc{solo}($p_i$): an agent solves $p_i$ alone, producing the baseline patch.}
\item \rev{\textsc{blind}($\{p_i\}$): agents solve their assigned tasks independently on $b$, without seeing the other agents' changes, and the resulting code diffs are merged.}
\item \rev{\textsc{informed}($p_{\pi(1)},\ldots,p_{\pi(N)}$): agents work in order $\pi$, each seeing the completed edits of earlier agents. This is the sequential oracle.}
\end{itemize}
\rev{We evaluate patches alone (\textsc{solo}) and together (\textsc{blind}) on every instance. On the synthetic tier, we also run the \textsc{informed} condition (\S\ref{sec:study}). The completed edit provides an ideal-visibility baseline for evaluating coordination.}
Following SWE-bench~\cite{swebench}, we grade code changes using hidden tests from the human PRs. Before grading, we discard any test-file edits made by the agent and apply the gold test patches. We then run the combined test set from all tasks in \emph{every} setting. Using different tests for individual and merged patches can falsely suggest interference (\S\ref{sec:study}).

\paragraph{The Interference Metric.}
Let $F_i$ be the tests that fail with patch $i$ alone, and let $F_{\mathrm{blind}}$ be the tests that fail after merging the patches. Both sets come from the same combined test suite. We define
\[
\Delta_{\mathrm{blind}} \;=\; \big|\,F_{\mathrm{blind}}\setminus \textstyle\bigcup_i F_i\,\big|,
\]
\rev{$\Delta_{\mathrm{blind}}$ counts failures that appear only when the patches are combined. A test that already fails with any individual patch is excluded. We reuse the exact patches from \textsc{solo} in the \textsc{blind} merge, so differences between agent runs cannot explain the result.

We report the count $\Delta_{\mathrm{blind}}$ and an indicator of whether any interference was detected, $I_{\mathrm{blind}}=\mathbf{1}[\Delta_{\mathrm{blind}}>0]$. For example, three newly failing tests give $\Delta_{\mathrm{blind}}=3$ and $I_{\mathrm{blind}}=1$; they need not represent three separate defects. Both measures depend on test coverage, and the count also depends on how tests are divided or duplicated. We therefore compare counts only with the same test set. A pair is \emph{observable}, and eligible for our interference analysis, only when each agent's patch passes its own task's tests during grading on the combined suite.}

\paragraph{\rev{Three Benchmark Tiers.}}
\rev{The \emph{synthetic} tier tests whether stale information causes interference by varying the number of changed interfaces while keeping the task and type of change fixed. The \emph{mined} tier uses reviewed PRs to test whether repository history contains examples of unresolved interference. The \emph{real-derived} tier applies scripted changes to real Django helpers to test whether the synthetic failure mechanism also occurs in existing code.

We also evaluate a communication condition, \textsc{comm}. The agent receives a short message describing the other worker's completed change. We generate this message from a template and the known transformation. This is an oracle message based on a completed change. We do not evaluate messages generated during ongoing work.}

\section{Dataset Construction}
\label{sec:construction}
We build most instances from Django, which has established tools for reproducible agent evaluation. We use the same deterministic construction procedure for SymPy, xarray, and seaborn.

\paragraph{Mining.} We collect merged PRs, keep those that change 1--12 runtime source files (excluding docs/CI/formatting), and form \emph{candidate pairs} that edit at least one common source file, an inexpensive way to find potentially related changes~\cite{chockchowwat2025chipmink}. Sharing a file does not guarantee interference; we test for it in the agent study.

\paragraph{Checking the Shared Base.} \rev{For each candidate pair, we choose the merge base of the two PR bases as the shared starting commit $b$. We then apply two filters, starting with the cheaper check. \emph{(i) Clean merge:} both code patches apply to $b$, and their composition produces no textual conflict.}
\rev{\emph{(ii) Base validity:} for each PR, its gold tests must fail on $b$ when only the test patch is applied and pass after adding the code patch. The gold test patch comprises the behavioral tests added or modified by the human PR.}

\rev{Checking the shared base matters because related PRs often start from different commits. One PR's base may already contain part of the other PR's solution, hiding the interaction we want to test. In one HTTP-header pair, for example, the later PR's base already contained a helper added by the earlier PR. Our check rejects such a base because the relevant tests pass before we apply the candidate fix. In a representative batch, invalid bases accounted for about one third of rejections.}

\paragraph{Yield.} From \emph{685} mined Django PRs we formed \emph{4{,}262} candidate pairs; deterministic validation across a 5-node cluster (320 cores) accepted \emph{417} as valid coupled instances. In a representative batch, ${\sim}36\%$ failed because the two changes \emph{textually} conflict, ${\sim}33\%$ failed base validity (drift), ${\sim}17\%$ did not apply on a common base, and ${\sim}14\%$ were accepted. We limit validation cost by running tests only on pairs that pass the structural and textual filters.

\begin{table}[t]
\centering\small
\caption{\rev{Benchmark summary across two model families and more than $4{,}000$ graded task solves. A validated coupled pair consists of two patches that apply and merge textually on a shared, base-valid commit; interference is measured separately by the agent study.}}
\label{tab:glance}
\begin{tabular}{lrrl}
\toprule
Tier & Instances & Runs & Role\\
\midrule
Synthetic (planted)        & $3{\times}4$ & $96$  & \rev{controlled changes}\\
Real mined, Django         & $417$        & $834$ & \rev{reviewed PRs}\\
Real mined, other repos    & $30$         & ---   & \rev{mining in other repos}\\
Disjoint controls          & $23$         & ---   & unrelated changes\\
Constructed real-derived   & $36$         & $216$ & \rev{real Django helpers}\\
\bottomrule
\end{tabular}
\\[2pt]
{\raggedright\footnotesize Synthetic: $3$ mechanisms $\times$ $4$ degrees, $8$ trials each ($96$ blind${+}$comm cells). Real-derived: $12$ real symbols $\times$ $3$ mechanisms $\times$ $3$ trials $=$ $108$ blind ${+}$ $108$ communication runs.\par}
\end{table}
\section{Preliminary Study}
\label{sec:study}
\rev{We evaluate more than $4{,}000$ graded task solves using two model families (Table~\ref{tab:glance}). The synthetic experiments measure interference as more interfaces change and test how much an oracle message can prevent. We then examine reviewed PR pairs for interference and apply the controlled changes to real Django helpers.}

\paragraph{Synthetic Tier.} \rev{Each synthetic instance adds $K{=}4$ helper functions to a Django checkout. The agent writes small wrappers that call these helpers. A scripted concurrent patch changes exactly $d$ helper interfaces, where $d\in\{0,\dots,4\}$ is the \emph{staleness degree}. It either renames helpers, adds a required argument, or changes the return type. The expected wrapper output stays the same. We run eight trials per setting. A blind agent uses the old interfaces, so we expect its wrappers to fail on the $d$ changed helpers after merging. An informed agent sees the new interfaces before writing its wrappers.}

\rev{Renames and required-argument changes matched the prediction. Blind patches failed on all $d$ changed helpers ($\Delta_{\mathrm{blind}}=d$, with zero variance). The informed agent had no failures. Because the wrapper tasks leave little room for alternative implementations, the result does not predict how failures scale in open-ended development. For return-type changes, the informed agent still averaged $0.13$ failures per run despite seeing the new interface.}

\begin{figure}[t]
\centering
\includegraphics[width=0.95\columnwidth]{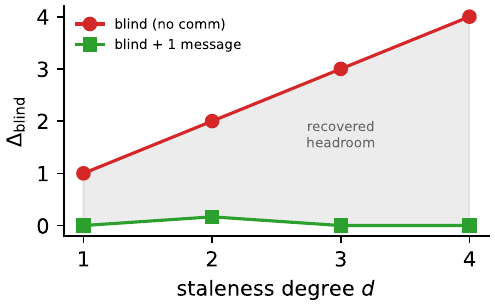}
\caption{\rev{Synthetic tier ($8$ trials per point, averaged over three mechanisms). Blind interference $\Delta_{\mathrm{blind}}$ increased with staleness degree $d$ (red). A ${\sim}130$-token oracle description of the finalized transformation reduced it to near zero (green), recovering $98\%$.}}
\Description{A line plot comparing blind interference with interference after an oracle message as the number of changed helper interfaces increases. The oracle message reduces the mean interference count from 2.50 to 0.04.}
\label{fig:comm}
\end{figure}

\paragraph{Synthetic Tier with an Oracle Message.} \rev{We next give the blind agent a message describing the scripted change, averaging about 130 tokens. The message states the new helper names, required arguments, or return types. With this information, the mean interference count fell from $2.50$ to $0.04$, a $98\%$ reduction (Fig.~\ref{fig:comm}). We generate the message from the scripted transformation, so the result assumes a complete and accurate description of the change.}

\paragraph{Mined PR Tier.} \rev{For the mined PR pairs, a GPT-5-class coding model solves each task in \textsc{solo} and \textsc{blind} conditions on the shared base. We run the model with mini-swe-agent~\cite{sweagent}, a minimal tool-calling harness, and with OpenHands~\cite{openhands}. Both use SWE-bench-style grading.}

\rev{Our initial grading procedure produced misleading interference counts for two reasons. First, agents sometimes edited test files, which could prevent the gold tests from being applied consistently. Second, we tested each individual patch only on its own task's tests, but tested merged patches on both tasks' tests. A patch could therefore break another task's test on its own, yet the failure would appear only in the merged evaluation. Both problems falsely increased $\Delta_{\mathrm{blind}}$ (Fig.~\ref{fig:pitfall}).}

\begin{figure}[t]
\centering
\footnotesize
\begin{tikzpicture}[
  cell/.style={draw,minimum width=12mm,minimum height=6.5mm,font=\small,align=center},
  pass/.style={cell,fill=green!12},
  fail/.style={cell,fill=red!13},
  na/.style={cell,fill=black!7},
  hdr/.style={font=\small\bfseries,align=center},
  lbl/.style={font=\small\itshape,align=right}]
\node[hdr] at (1.1,0.62) {\textsc{solo}\,A};
\node[hdr] at (2.45,0.62) {\textsc{solo}\,B};
\node[hdr] at (3.8,0.62) {\textsc{blind}};
\node[hdr] at (5.35,0.62) {$\Delta_{\mathrm{blind}}$};
\node[lbl] at (-0.35,0) {naive};
\node[pass] at (1.1,0) {pass};
\node[na]   at (2.45,0) {not run};
\node[fail] at (3.8,0) {fail};
\node[font=\small] at (5.35,0) {$+1$~\emph{(false)}};
\node[lbl] at (-0.35,-0.78) {corrected};
\node[pass] at (1.1,-0.78) {pass};
\node[fail] at (2.45,-0.78) {fail};
\node[fail] at (3.8,-0.78) {fail};
\node[font=\small] at (5.35,-0.78) {$0$~\emph{(excluded)}};
\end{tikzpicture}
\caption{\rev{Using different test sets can create false interference. Patch~B fails this test on its own, but the initial grading procedure does not run it in \textsc{solo}\,B. Running the same tests in all conditions reveals the existing failure and excludes it from $\Delta_{\mathrm{blind}}$.}}
\Description{Two rows compare task-specific and combined-test grading. Task-specific grading misses a failure in patch B alone and counts it as interference after merging. Combined-test grading detects the existing failure and excludes it.}
\label{fig:pitfall}
\end{figure}

\rev{After removing agent test edits and running the combined test set in every condition, nearly all apparent interference disappeared. Of 834 runs on 417 validated Django pairs, only one had $\Delta_{\mathrm{blind}}>0$, with one newly failing test. The reviewed human patches also merged without gold-test failures across the examined pairs. These results suggest that merged PR history is a poor source of examples of unresolved parallel changes: development and review may have already made the patches compatible.}

\rev{GPT-5.5 reproduced the synthetic result on newly added helper names, reducing the likelihood that the result depends on memorized code. On the most strongly coupled mined pairs, it showed no interference in 72 runs. It did fail on the constructed Django tasks described next. The difference is consistent with the mined patches having been made compatible during development and review.}

\paragraph{Constructed Tasks on Real Django Helpers.}
\rev{We apply three interface transformations to 12 Django helpers to construct 36 instances (Appendix~\ref{app:real}). Each helper is linked to a validated PR pair that illustrates a dependency through a shared symbol. One patch makes a scripted interface change; an agent writes the other patch, which uses the helper.

In the blind condition, GPT-5.5 produced interference in 105/108 runs (97\%), with failures on all 12 helpers despite clean textual merges. With the oracle message, 89/108 runs (82\%) were recovered. Recovery reached 93\% among message-conditioned patches that merged cleanly. These rates apply to the constructed Django tasks.}

\rev{The mining pipeline also works beyond Django. With the same pytest-based validation harness, it finds 11 valid pairs in SymPy, 12 in xarray, and 7 in seaborn. Textual conflicts and incompatible base commits account for most rejections. In practice, construction works best with PRs close in time and tests selected from the PR's own changes. Running entire modified test files can introduce unrelated or flaky cases.}

\paragraph{Threats to Validity.}
The main measurement risk is counting an individual patch's failure as interference. We address this by discarding agent test edits, running the same combined test set in every condition, and requiring each patch to solve its own task in isolation. As in SWE-bench~\cite{swebench}, we rely on the gold tests to define correct behavior. We also check that those tests fail on the shared base before applying the human fix.

\rev{Historical PRs have been through review, while the constructed tasks deliberately introduce breaking changes. The latter use one scripted patch and constrain the agent to call selected helpers. Our agent experiments are also limited to Django. The results establish the failure mechanism and demonstrate it on 12 real helpers, but do not estimate its frequency or severity in everyday parallel development.

The communication results assume complete descriptions of finalized changes; recovery with messages generated during parallel work remains untested. Finally, tests can miss failures, and several tests can detect the same defect. We therefore interpret interference counts only for a shared test set.}

\section{Ongoing Work and Open Problems}
\label{sec:ongoing}
\rev{The experiments leave open how often interference occurs before review, how agents should communicate during parallel work, and how the results extend to more than two tasks.}

\paragraph{Parallel Work Before Review.}
\rev{To measure how often interference occurs in practice, we need execution records from parallel agents before review or repair. These records should preserve a shared starting commit, independently generated patches, and intermediate states~\cite{li2024kishu}. These records would let us compare information gaps in systems with isolated workers, asynchronous updates, explicit orchestration, or shared state.}

\paragraph{\rev{From Oracle Visibility to Online Communication.}}
\rev{The oracle message reduces synthetic interference by $98\%$. In a running system, however, agents may only be able to share their plans, an unfinished patch, or an automatically generated summary. These options differ in completeness, delay, and privacy~\cite{xia2026slotguard}. A comparison with affected-symbol lists and completed edits, using the same token budget, would show how message content and timing affect recovery.}

\paragraph{Scale.}
\rev{Our experiments focus on pairs of tasks. With $N>2$ tasks, a failure may require three or more changes to occur together, even if every pair works correctly. Larger task sets also create more possible dependencies. Extending the benchmark would test whether results for pairs predict reliability for larger groups, and whether we can estimate coordination headroom for a given set of tasks before agents start work.}

\section{Conclusion}
\rev{Agents working from the same codebase can produce patches that pass alone but fail together, even when the merge has no textual conflicts. In \textsc{stale}, we test the same patches alone and together using a common test set. Comparing blind agents with agents that see the completed concurrent change estimates how much coordination could help.}

\rev{Synthetic tasks isolate failures caused by stale interfaces. Mined PR pairs suggest that reviewed history rarely retains unresolved interactions. Constructed tasks reproduce the mechanism on 12 Django helpers, with 97\% interference and 82\% recovery using an oracle message. These constructed rates do not measure real-world prevalence. The next step is to study parallel work before review, across more repositories, with messages that agents can generate during execution.}

\begin{acks}
NSF grants \#2103794, \#2312991, \#2551201, \#2440498, and \#2312561 support this work, along with DAPLab corporate support in the form of funding and/or compute from Amazon, IntellectAI, Infosys, Tidalwave, Veris, Shopify, Microsoft, Thinking Machines, Dandy, Perplexity, and Daytona. This work is also supported by the Advanced Cyberinfrastructure Coordination Ecosystem: Services \& Support (ACCESS) program, which is supported by National Science Foundation grants \#2138259, \#2138286, \#2138307, \#2137603, and \#2138296. The views and conclusions presented here are those of the authors and should not be interpreted as representing the official positions of the funding or supporting organizations.
\end{acks}

\balance
\appendix

\section{Related Work}
\paragraph{Agent Benchmarks and Scaffolds.}
SWE-bench~\cite{swebench} evaluates coding agents on real GitHub issues using reproducible tests. SWE-agent~\cite{sweagent} provides the agent--computer interface used in our experiments. These tools evaluate one task at a time. We use the same test-based approach to measure whether changes developed in parallel still work when combined.

\paragraph{Multi-Agent Software Development.}
AutoGen~\cite{autogen}, MetaGPT~\cite{metagpt}, and ChatDev~\cite{chatdev} coordinate LLM agents through assigned roles and communication protocols. We measure failures caused by stale information and compare them with an oracle condition in which agents see completed concurrent changes. Evaluating messages based on plans or unfinished work remains an open question.

\paragraph{Merging and Collaboration Conflicts.}
Software-merging research has studied both textual and semantic conflicts~\cite{mens,sousa}. Some tools detect conflicts by merging developer branches before developers commit their work~\cite{brun}; transactional approaches use isolation to avoid conflicts before commit~\cite{chockchowwat2023transactional}. We study patches produced by independent coding agents and use hidden tests to detect behavioral failures after a clean textual merge.

\section{A Minimal Worked Instance}
\label{app:example}
Consider a synthetic module with $K{=}4$ helper functions that put brackets around values. Task~A adds wrappers \texttt{f1..f4} that call these helpers. Task~B renames \texttt{wrap}$k$ to \texttt{bracket}$k$ without changing the returned values. The staleness degree $d$ is the number of helpers B renames. For the first helper, A writes:
\begin{lstlisting}[language=Python]
def wrap1(value):              # shared helper; Task A calls it
    return "[" + str(value) + "]"
def f1(values):                # Task A, written against wrap1
    return ",".join(wrap1(v) for v in values)
\end{lstlisting}
With Task~A alone, \texttt{f1([1,2])} returns \texttt{"[1],[2]"}. Task~B preserves the helper's behavior. The patches edit different regions, so \texttt{git} merges them without a conflict. After the merge, however, A's wrapper calls a name that B has removed:
\begin{lstlisting}[language=Python]
>>> f1([1, 2])
NameError: name 'wrap1' is not defined
\end{lstlisting}
An informed agent sees B's rename before writing the wrapper and calls \texttt{bracket1}. In \textsc{comm}, the benchmark instead sends a message describing the completed edit: ``\texttt{wrap1} was renamed to \texttt{bracket1}.'' The agent uses the new name and the combined code passes.

\section{A Real-Derived Staleness Tier}
\label{app:real}
We construct this tier using existing Django helpers to test the failure mechanism beyond the helpers added for the synthetic experiments. GPT-5.5 writes a consumer that calls a helper, and a scripted patch changes that helper's interface.

\paragraph{Construction.} We select 12 helpers that can be unit-tested from Django's text, HTML, HTTP, and encoding modules. Each helper is linked to a validated PR pair in which one patch modifies a symbol used by the other ($W_A\cap R_B>0$). The historical pair supplies an example of the dependency; we construct a breaking version on a helper that we can test in isolation.

For example, we link \texttt{phone2numeric} to PRs~18361 and~20309. The historical dependency involves different symbols: PR~18361 changes the signature of \texttt{as\_sql}, and PR~20309 adds an \texttt{as\_sqlite} method that calls it. Those reviewed changes work together because the signature change is additive. Our constructed task makes a breaking change to a helper while asking an agent to add a feature that calls it.

We apply three transformations: \emph{rename} removes the old helper name; \emph{signature} adds a required keyword-only argument; and \emph{rettype} returns an object whose \texttt{.text} field contains the original result. Applying these transformations to 12 helpers produces 36 instances.

\paragraph{Worked Example.} Task~A renames \texttt{phone2numeric}. Task~B asks for a new function, \texttt{phone2numeric\_\allowbreak clean}. It strips spaces and hyphens from its input, then uses the existing helper to convert letters to digits. Working on the base commit without seeing A's change, GPT-5.5 writes:
\begin{lstlisting}[language=Python]
def phone2numeric_clean(phone):
    cleaned = str(phone).replace(" ", "").replace("-", "")
    return phone2numeric(cleaned)   # calls the OLD name
\end{lstlisting}
B passes alone, and the patches merge without a textual conflict. The combined code fails because B still calls the old helper name:
\begin{lstlisting}
NameError: name 'phone2numeric' is not defined. Did you mean: 'phone_to_numeric'?
\end{lstlisting}
In \textsc{comm}, we generate a message describing Task~A's completed rename. The agent calls the new name, and the combined code passes.

\begin{table}[t]
\centering\small
\caption{\rev{Real-derived Django tier: GPT-5.5, $3$ scripted mechanisms $\times$ $3$ trials per
symbol ($9$ runs each). \textsc{blind} denotes silent interference; \textsc{comm} receives an
oracle description of the finalized scripted edit. Across these constructed instances, blind
interference occurs in $105/108$ runs ($97\%$), and the oracle message recovers $89/108$ ($82\%$;
$93\%$ among cleanly merging cases).}}
\label{tab:realderived}
\begin{tabular}{llccc}
\toprule
Helper & Module & Anchor pair & \textsc{blind} & \textsc{comm}\\
\midrule
\texttt{phone2numeric}    & text     & 18361,20309 & 8/9 & 9/9\\
\texttt{slugify}          & text     & 17554,19874 & 9/9 & 6/9\\
\texttt{get\_text\_list}  & text     & 18361,19073 & 9/9 & 9/9\\
\texttt{capfirst}         & text     & 19244,19445 & 9/9 & 8/9\\
\texttt{get\_valid\_filename} & text & 19955,20458 & 9/9 & 9/9\\
\texttt{normalize\_newlines} & text  & 16703,19478 & 9/9 & 9/9\\
\texttt{escape}           & html     & 19057,20309 & 9/9 & 4/9\\
\texttt{strip\_tags}      & html     & 19459,20309 & 9/9 & 8/9\\
\texttt{int\_to\_base36}  & http     & 20587,21092 & 9/9 & 9/9\\
\texttt{urlencode}        & http     & 19919,21531 & 8/9 & 3/9\\
\texttt{iri\_to\_uri}     & encoding & 19040,19445 & 8/9 & 8/9\\
\texttt{base36\_to\_int}  & http     & 19555,20309 & 9/9 & 7/9\\
\midrule
\multicolumn{3}{l}{\emph{total}} & \textbf{105/108} & \textbf{89/108}\\
\bottomrule
\end{tabular}
\end{table}

\paragraph{Results.} Across the 12 helpers, 105 of 108 blind runs show interference (Table~\ref{tab:realderived}). In each counted run, the consumer passes alone, merges without textual conflicts, and fails the hidden tests after the scripted transformation. We observe this failure on all 12 helpers.

With the oracle message, 89/108 runs (82\%) recover. Among message-conditioned patches that merge without textual conflicts, recovery reaches 93\%. All 36 return-type runs recover, and recovery is also high for renames. For required-argument changes, the agent sometimes edits the helper definition itself, creating a textual conflict.

\paragraph{Scope.} These tasks reproduce the stale-interface failure using real Django helpers and dependency patterns found in PR history. We script the breaking patch and give the communication condition a complete description of it. Estimating how often unresolved dependencies occur in practice requires records of parallel agent work before review.

\bibliographystyle{ACM-Reference-Format}
\bibliography{references}
\end{document}